\RequirePackage{fix-cm}
\documentclass{svjour3}                     
\smartqed  
\usepackage[numbers]{natbib}
\usepackage{graphicx}
\usepackage{ulem}
\usepackage{dsfont}
\usepackage{bbm}
\usepackage{amsfonts}
\usepackage{color}
\usepackage{array}
\usepackage[misc]{ifsym}
\usepackage{multicol}
\usepackage{setspace}
\usepackage{amsmath}
\usepackage{amssymb}
\usepackage[pdftex,
            pdfstartview=FitH,
            CJKbookmarks=true,
            bookmarksnumbered=true,
            bookmarksopen=true,
            linkcolor=green,
            anchorcolor=green,
            citecolor=green
            ]{hyperref}
\begin{document}

\title{Multi-Modal Gated Recurrent Units for Image Description
}


\author{Xuelong Li \and Aihong Yuan \and Xiaoqiang Lu
}


\institute{Xuelong Li \at
              Center for OPTical IMagery Analysis and Learning (OPTIMAL), Xi'an Institute of Optics and Precision Mechanics, Chinese Academy of Sciences, Xi'an 710119, Shaanxi, P. R. China.
              \at University of Chinese Academy of Sciences, 19A Yuquanlu, Beijing, 100049, P. R. China.\\
              \email{xuelong\_li@opt.ac.cn}
           \and
           Aihong Yuan \at
              Center for OPTical IMagery Analysis and Learning (OPTIMAL), Xi'an Institute of Optics and Precision Mechanics, Chinese Academy of Sciences, Xi'an 710119, Shaanxi, P. R. China.
              \at University of Chinese Academy of Sciences, 19A Yuquanlu, Beijing, 100049, P. R. China.\\
           \email{ahyuan@opt.ac.cn}
           \and
           Xiaoqiang Lu \Letter \at
              Center for OPTical IMagery Analysis and Learning (OPTIMAL), Xi'an Institute of Optics and Precision Mechanics, Chinese Academy of Sciences, Xi'an 710119, Shaanxi, P. R. China.\\
           \email{luxq666666@gmail.com}
           \and
           2018 Springer. Personal use of this material is permitted. Permission from Springer must be obtained for all other uses, in any current or future media, including reprinting/republishing this material for advertising or promotional purposes, creating new collective works, for resale or redistribution to servers or lists, or reuse of any copyrighted component of this work in other works.
}

\date{Received: 21 May 2017 / Revised: 4 January 2018 / Accepted: 1 March 2018}

\maketitle

\begin{abstract}
Using a natural language sentence to describe the content of an image is a challenging but very important task. It is challenging because a description must not only capture objects contained in the image and the relationships among them, but also be relevant and grammatically correct. In this paper a multi-modal embedding model based on \textsl{gated recurrent units} (GRU) which can generate variable-length description for a given image. In the training step, we apply the \textsl{convolutional neural network} (CNN) to extract the image feature. Then the feature is imported into the multi-modal GRU as well as the corresponding sentence representations. The multi-modal GRU learns the inter-modal relations between image and sentence. And in the testing step, when an image is imported to our multi-modal GRU model, a sentence which describes the image content is generated. The experimental results demonstrate that our multi-modal GRU model obtains the state-of-the-art performance on Flickr8K, Flickr30K and MS COCO datasets.
\end{abstract}

\section{Introduction}
\label{intro}
Image description aims to use a natural language sentence to describe the content of the given image. Image description requires detailed understanding of an image: we need not only detect objects contained in the image, but also find their relationships and their attributes. It is one of the ultimate goals of \textsl{artificial intelligence} (AI), \textsl{computer vision} (CV), \textsl{natural language processing} (NLP) and \textsl{matching learning} (ML) \cite{yan2015dcca}.
 This task is ambitious and has many important applications, such as image retrieval (in fact, image retrieval with natural language as a query is more natural than image as a query), early childhood education and visually impaired people assistance.

\begin{figure*}
  \centering
  \includegraphics[width=0.90\linewidth]{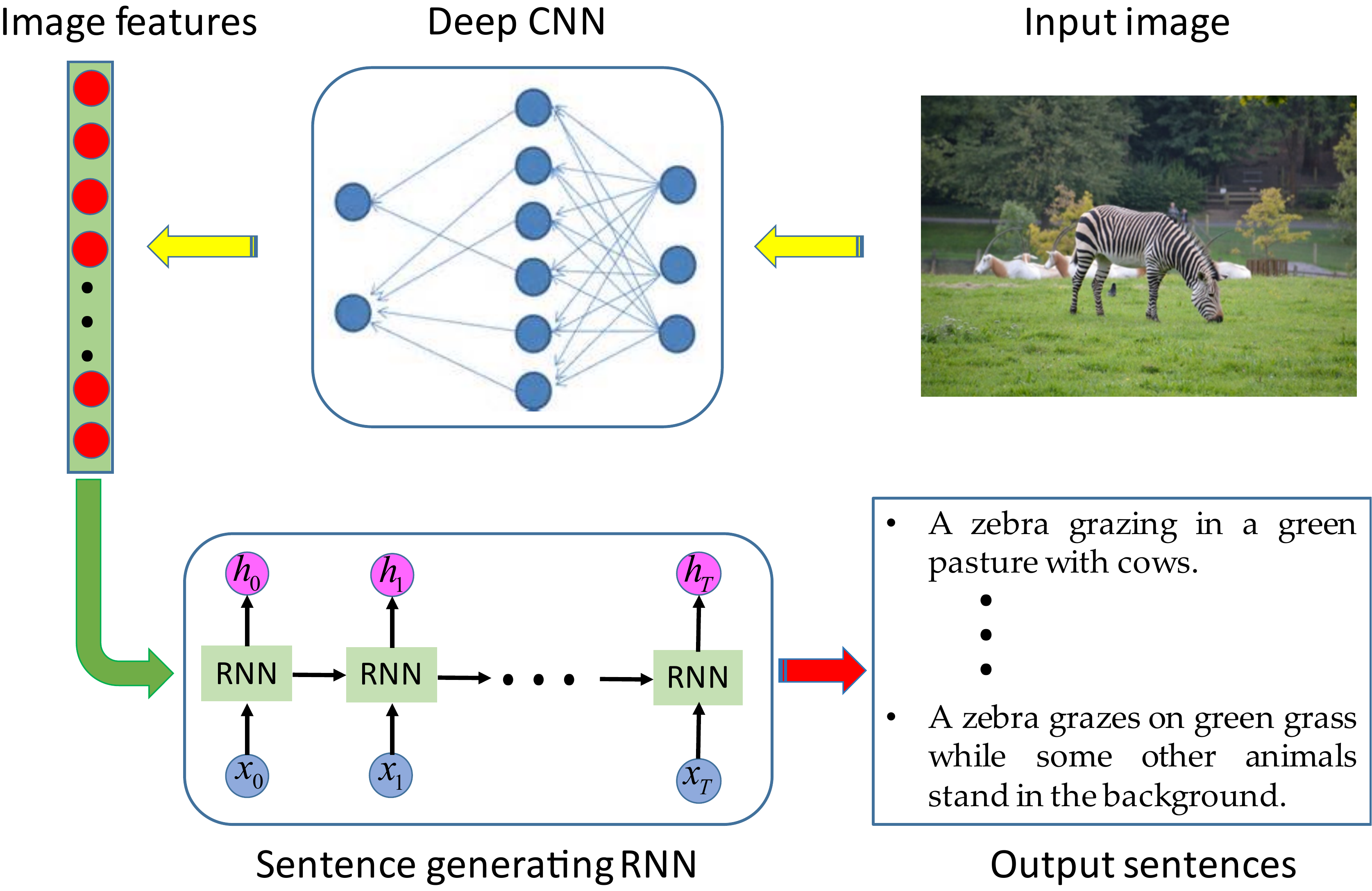}\\
  \caption{An generative ``CNN + RNN'' architecture for image description generation. A CNN model for image feature representation and a RNN model for sentence generating.}
  \label{CNN-RNN model_fig}
\end{figure*}

\par Many pioneers have done research on image description and gained some progress. Approaches for image description task can be roughly categorized three categories: 1) template-based approaches; 2) retrieval-based approaches; and 3) \textsl{multi-modal neural networks-based} (MMNN-based) approaches. Template-based approaches are mainly composed two important parts: hard-coded language templates and visual concepts\cite{everingham2010the,kulkarni2013baby}. Because the language templates and visual concepts are fixed, sentences generated by these models are less of variety. Retrieval-based approaches retrieve similar captioned images and generate new descriptions by retrieving a similar sentence from the training dataset \cite{hodosh2013framing,socher2014dt-rnn}. These approaches project image features and sentence features into a common space, which is used for ranking image captions or for image search. However, they cannot generate novel sentence from the embedding. In recent years, with the development of deep learning, the \textsl{convolutional neural network} (CNN) and \textsl{recurrent neural network} (RNN) have gotten a lot exploits in CV and NLP fields \cite{szegedy2015googlenet,simonyan2014vggnet,krizhevsky2012alexnet,cho2014gru,chung2014empirical,sutskever2014sequence}. Under such circumstances, MMNN-based approaches become the most effective and popular approaches. Their core idea is that regarding image-to-sentence task as a \textsl{machine translation} (MT) problem (see Fig. \ref{CNN-RNN model_fig}). In other words, when an image is imported into their models, it will be ``translated" to an English sentence. They use CNN to extract image features and RNN to generate sentences. Although these methods have made great progress, they have some shortages. Some of them use traditional RNN to generate sentences \cite{mao2014m-rnn,karpathy2015devs,chen2015lrvr}, but traditional RNN has a fatal flaw---“vanishing gradients” or “exploding gradients”---which makes their models hard to train \cite{chung2015gated}. Others use LSTM as sentence generator \cite{kiros2014mnlm,donahue2015lrcn,vinyals2015nic}. LSTM model can solve the “exploding gradient” and “vanishing gradient” phenomenon in traditional RNN, but the structure of LSTM is more intricate which leads to models have more parameters and need more time to train \cite{cho2014gru}.

\par To solve aforementioned problems, we propose a multi-modal \textsl{gated recurrent unit} (GRU) model. It contains two important parts: an ``encoder'' CNN and a ``decoder'' RNN. The ``encoder'' CNN encodes input images into feature vectors. Then these feature vectors are transformed into a fixed-length vector (see Section-\ref{Image Representation}) and imported into the ``decoder'' RNN. At last, the ``decoder'' RNN generates sentence description for the images.
\par Fig. \ref{m-GRU model_fig} shows the diagram of our multi-modal GRU network generative model. VGG-16 network \cite{simonyan2014vggnet} is used as the ``encoder''. We get rid of the last FC-1000 and soft-max layers, and the last FC-4096 layer's output is used as the feature representation of an image. On the contrary to \cite{mao2014m-rnn,kiros2014mnlm,donahue2015lrcn,vinyals2015nic,karpathy2015devs,chen2015lrvr}, we use GRU as sentence generator. In the training stage, the aforementioned image features and the corresponding sentence representations are imported into the multi-modal GRU which learns the inter-modal relation between images and sentences. All parameters are learned by maximizing the likelihood function with the backpropagated algorithm. In the predicting stage, image features are imported into the multi-modal GRU, and the corresponding description sentence is generated by the multi-modal GRU.
\par Our model can generate variable-sized descriptions for input images. In the experiments, we train and test our model on three benchmark datasets---Flickr8K \cite{rashtchian2010flickr8k}, Flickr30K \cite{young2014flickr30k} and MS COCO \cite{lin2014coco}. The experimental results show that our multi-modal GRU model gains the state-of-the-art performance on the benchmark datasets.
\par Our core contributions are listed as follows:

\begin{itemize}
\item We present an end-to-end neural model for the image-to-sentence problem. Our model can be fully trained with the \textsl{stochastic gradient descent} (SGD) method. This model is simple but very effective and general.
\item The GRU model is first applied to the image-to-sentence problem and the model has finished the task excellently. The GRU can not only solve the vanishing gradients and exploding gradients problems like LSTM, but also has a much simpler architecture than the LSTM which makes our model much easier to train.
\item We have proposed two variations of our multi-modal GRU model architecture: 1-layer model and 2-layer model. We show that both the two variations can complete the image description very well and by the contrast, the 2-hidden-layer model performs better than the 1-hidden layer model if the training dataset is large enough.
\end{itemize}

\par The rest of the paper is organized as follows. In Section-\ref{Related Work}, some precious works related to ours are briefly introduced. Then our model is described in Section-\ref{our model}. The competitors and comparing experimental results are shown in Section-\ref{Experiments}. At last, Section-\ref{conclusion} makes a briefly conclusion for this paper.
\section{Related Work} \label{Related Work}
\textbf{Deep neural network for CV.} In recent years, deep neural network has enjoyed a great success in the field of image representation and video processing, even in the remote sensing image recognition \cite{DBLP:journals/tgrs/YaoHCQ016,DBLP:journals/tip/YaoHZN17,DBLP:journals/pieee/ChengHL17,DBLP:journals/ijcv/ZhangHLWL16,DBLP:journals/tnn/ZhangHHS16,cheng2017remote}.
Many deep \textsl{convolutional neural network} (CNN) have been designed to solve CV problems. Researches have shown that  CNN can learn representative and discriminative features in a hierarchical manner from the image data \cite{szegedy2015googlenet,simonyan2014vggnet,krizhevsky2012alexnet}. A very important and typical CNN architecture called LeNet was first proposed by LeCun in 1998 \cite{lecun1998lenet}. The CNN was previously trained for object recognition tasks. After Hinton published a paper about deep neural network on Science in 2006 \cite{hinton2006reducing}, deep learning has gained a rushed development.
With the \textsl{ImageNet Large Scale Visual Recognition Competition} (ILSVRC) \cite{russakovsky2015imagenet} capturing a lot of attention, some typical CNN models such as AlexNet \cite{krizhevsky2012alexnet} and GoogLeNet \cite{szegedy2015googlenet} have been designed. The AlexNet has gotten the first place in ILSVRC 2012 and the error ratio is 15\%. The GoogLeNet's depth is much deeper but the parameter is much less than the AlexNet and this structure has taken the first place in ILSVCR 2014 with the error ratio 6.66\%.
In 2014, Simonyan \& Zisserman \cite{simonyan2014vggnet} proposed a CNN with 16-19 layers which is called VGG-Net. Their main contribution is investigating how the CNN depth affects the accuracy in the large-scale image recognition setting.
The VGG-Net has been used for object recognition and image classification tasks and shows the state-of-the-art performance on image representation. So the VGG-Net has been exploited in many studies. In our image description task, we use the VGG-16 as an ``encoder" which encodes an RGB input image into a 4096-dimension vector.
\par \textbf{Deep neural network for MT.} Recently, many works showed that deep neural network can be successfully used to solve lots of problems in NLP, such as language model \cite{mikolov2010recurrent,mikolov2011strategies,mnih2007three,mikolov2013efficient,bengio2003neural}, paraphrase detection \cite{kiros2015skip,socher2011dynamic,dolan2004unsupervised}, word embedding extraction and so on. In the field of MT, deep neural network also showed nice performance and become a mainstream method.
In the conventional MT system, the neural network refers to as an RNN Encoder-Decoder which consists of two RNN \cite{cho2014gru,chung2014empirical,chung2015gated}. One acts as an encoder which maps a variable-length source sentence to a fixed-length vector. And the other acts as a decoder which decodes the vector produced by the encoder into a variable-length target sentence.
After unfolding the structure, RNN can be viewed as a type of feedforward neural network with shared transitional weights.
When RNN trained with \textsl{Backpropagation Through Time} (BPTT) \cite{werbos1990backpropagation,fairbank2013equivalence}, there exists some difficulties in learning long-term dependency which is due to the so-called vanishing and exploding gradient problems \cite{greff2015lstm,chung2015gated,hochreiter1997lstm}. To overcome these difficulties, some gated RNNs (\textsl{e.g.} LSTM \cite{hochreiter1997lstm} and GRU \cite{cho2014gru}) have been proposed. Compared with GRU, the structure of LSTM is more complex and has more parameters because every LSTM unit has three gates while GRU unit has only two gates. Taking these factors into account, we choose the GRU as the ``decoder".
\par \textbf{Multi-modal representation learning.} Image description is a multi-modal representation learning problem between images and text. Image and text are two different modalities, but they can represent the same information or complement each other. Data in different modalities can be utilized to discover the underlying latent correlation between data objects \cite{DBLP:journals/ijon/DingZGLZH17,DBLP:journals/tmm/ZhaoYGJD17,DBLP:journals/sigpro/ZhaoYZWL15,DBLP:journals/tip/DingGZG16,DBLP:journals/tip/GuoDHG17}.
\textsl{Canonical Correlation Analysis} (CCA) \cite{jin2015c++,gatignon2014canonical,hardoon2004canonical} is a typical method for multi-modal representation learning, which tends to map the multi-modal data into a common space and then minimizes the distance between the two modalities of data if they are correlated, otherwise maximizes it. Another similar method is \textsl{Latent Dirichlet Allocation} (LDA) \cite{blei2003latent,mei2015security} which attempts to model the correlation between multi-modal data. In recent years, deep architectures have also been conducted to learn the multi-modal representation \cite{ma2015multimodal,chen2014lrvr}. These models map the image representation and sentence representation into a common multi-modal embedding space. Multi-modal \textsl{Deep Boltzmann Machines} (DBMs) \cite{srivastava2012m-dbm,salakhutdinov2009deep} is a typical deep architecture used for multi-modal representation. Different specific DBMs extract features of different modal data. Then these features are projected into a multi-modal embedding space and jointly represented to learn the correlation between these modalities. This method is very useful and convenient and our multi-modal embedding model is similar with this model while there are some difference between them. Our multi-modal embedding model explicits the deep architecture with GRU.

\par \textbf{Generating sentence descriptions for images.}
There are mainly two categories of methods for this tasks. The first category is retrieval-based methods \cite{hodosh2013framing,socher2014dt-rnn,kuznetsova2014treetalk,jia2011learning,farhadi2010every} which retrieve similar captioned images and generate new descriptions by retrieving a similar sentence from a image-description dataset.
Another typical category is multi-modal neural network based methods \cite{DBLP:conf/aaai/ChenDZCLH17,DBLP:journals/corr/ZhangSLXGYH17,DBLP:journals/corr/AndersonHBTJGZ17,yan2015dcca,ma2015multimodal,mao2014m-rnn,kiros2014mnlm,donahue2015lrcn,vinyals2015nic,karpathy2015devs,chen2015lrvr}. These methods based off of multi-modal embedding models, generated sentence in a word-by-word manner and conditioned on image representation which is the output from a deep convolutional network.
Our method falls into this category and our multi-modal embedding model is a recurrent network. Our model is trained to maximize the likelihood of the target sentence conditioned on the given training image. Some previous works seem closely related to our method such as m-RNN \cite{mao2014m-rnn}, Google-NIC \cite{vinyals2015nic}, LRCN \cite{donahue2015lrcn} and so on. However, these methods use the traditional RNN or LSTM as the decoder which is different from us.

\section{Model Architecture} \label{our model}

\begin{figure}
  \centering
  \includegraphics[width=0.7\linewidth]{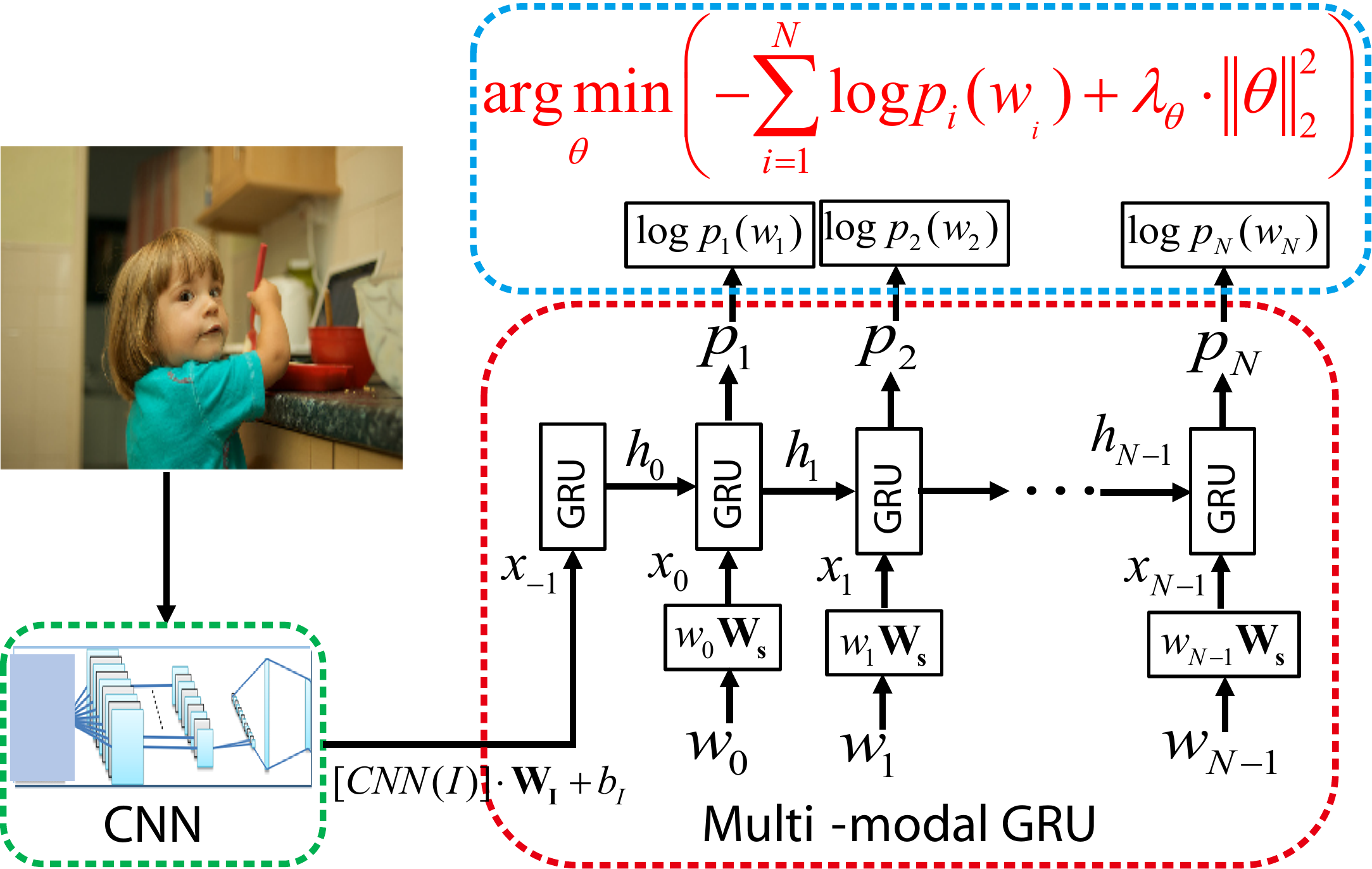}\\
  \caption{The diagram of our multi-modal GRU network generative model. An image is input into a CNN model and the feature of this image is extracted by the CNN network, represented as a feature vector. Then, the feature vector is fed into the multi-modal GRU network as the first ``word'' vector. Conditioned on the previous words and image, one word is generated at each time step.}
  \label{m-GRU model_fig}
\end{figure}

  Fig. \ref{m-GRU model_fig} shows the architecture of our multi-modal GRU model.
  During training procedure, CNN is used to extract the image feature. Then the feature is imported into the multi-modal GRU network with the corresponding sentence representations. The multi-modal GRU network learns the intermodal relation between image and sentence. Utilizing the large image-description datasets, we train the whole parameters of our model through maximizing the likelihood of the target description sentence which is given for the training image.
  In this section, we introduce every part of the model in detail.

\subsection{Gated Recurrent Unit} \label{GRU}

\textsl{Gated Recurrent Unit} (GRU) is first proposed recently by K. Cho \textsl{et al.} \cite{cho2014gru} and used in MT. The GRU model is  motivated by the LSTM unit and they have the same purposes---remembering the specific feature in the input stream for a long series of steps and avoiding vanishing gradients by gated architecture. If you want to know more about the differences and similarities between GRU and LSTM, you can refer to \cite{chung2014empirical,karpathy2015visualizing}.
GRU is only proposed two years long and increasing researchers have poured attention to this model. GRU is one of the most important parts of our algorithm, so in this section, we want to describe the GRU model at first.

\par
\begin{figure}
  \centering
  \includegraphics[width=0.6\linewidth]{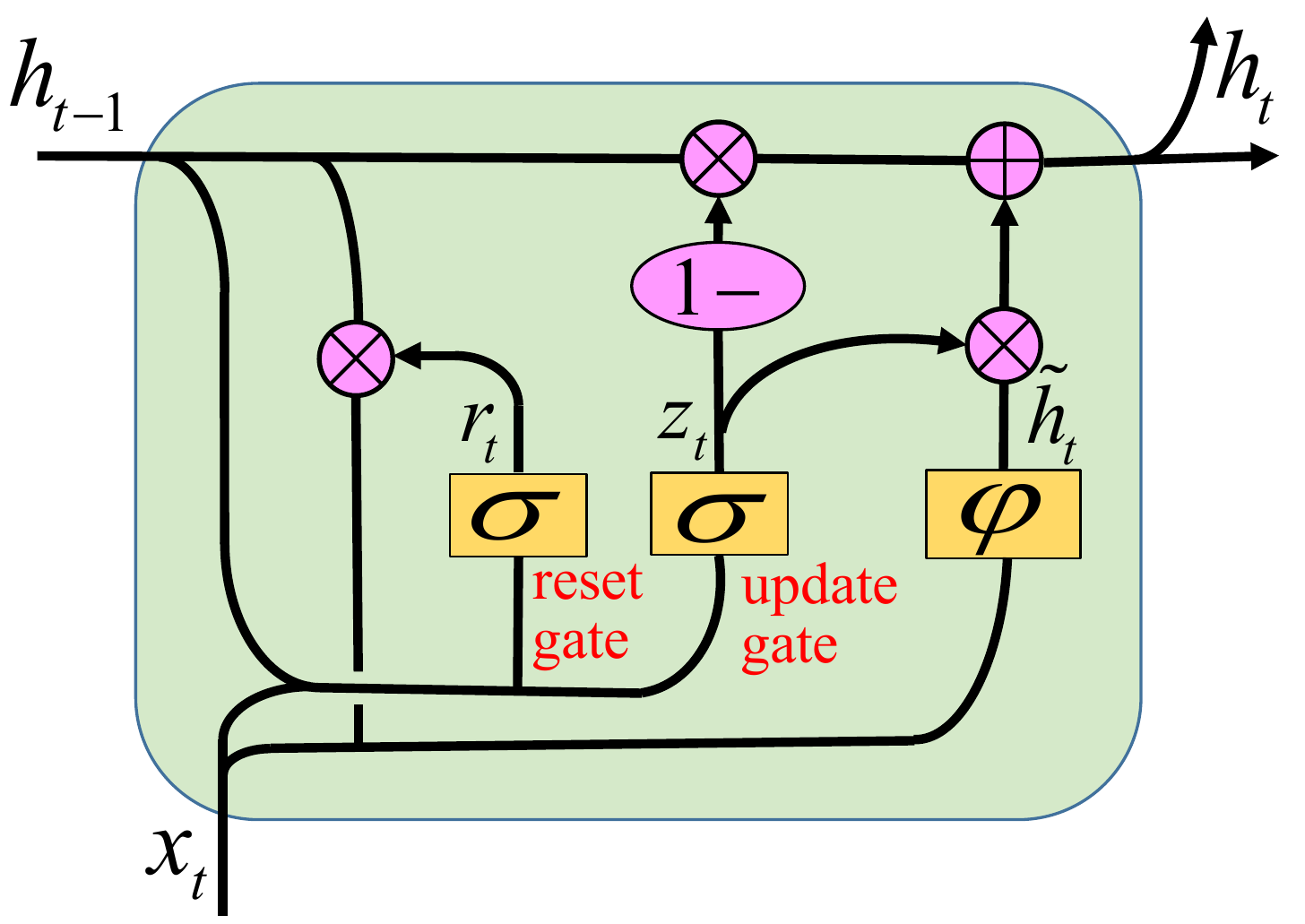}\\
  \caption{A diagram of a GRU block. The reset gate $r_{t}$ controls wether ignoring the previous hidden state $h_{t-1}$ or not. The update gate $z_{t}$ decides whether the hidden state $h_t$ is to be updated with a new hidden state $\tilde h_{t}$. Eqs. (\ref{reset gate})-(\ref{hidden}) show the detail of GRU block.}
  \label{GRU_fig}
\end{figure}

 Fig. \ref{GRU_fig} shows the architecture of GRU block. It features two gates (reset gate $r_{t}$, update gate $z_{t}$), a candidate activation ($\tilde h_{t}$) and a block output ($h_t$). The gates and date update are defined as follows:
\begin{equation}\label{reset gate}
  {\rm{{{\bf{R}}_{\bf{t}}} = \sigma ({{\bf{X}}_{\bf{t}}}\cdot{{\bf{W}}_{\bf{r}}} + {{\bf{H}}_{{\bf{t - 1}}}}\cdot{{\bf{U}}_{\bf{r}}} + {{\bf{b}}_{\bf{r}}})}},
\end{equation}
\begin{equation}\label{update gate}
  {\rm{{{\bf{Z}}_{\bf{t}}} = \sigma ({{\bf{X}}_{\bf{t}}}\cdot{{\bf{W}}_{\bf{z}}} + {{\bf{H}}_{{\bf{t - 1}}}}\cdot{{\bf{U}}_{\bf{z}}} + {{\bf{b}}_{\bf{z}}})}},
\end{equation}
\begin{equation}\label{1}
  {\rm{\bf{\tilde H}}_{\bf{t}}} = \varphi [{{\bf{X}}_{\bf{t}}}\cdot{{\bf{W}}_{\bf{h}}} + ({{\bf{R}}_{\bf{t}}} \odot {{\bf{H}}_{{\bf{t - 1}}}})\cdot{{\bf{U}}_{\bf{h}}} + {{\bf{b}}_{\bf{h}}}],
\end{equation}
\begin{equation}\label{hidden}
  {\rm{{{\bf{H}}_{\bf{t}}} = ({\bf{1}} - {{\bf{Z}}_{\bf{t}}}) \odot {{\bf{H}}_{{\bf{t - 1}}}} + {{\bf{Z}}_{\bf{t}}} \odot {{\bf{\tilde H}}_{\bf{t}}}}},
\end{equation}
where $\bf{X}_{\bf{t}}$ donates training instances (images and sentences representation of the training set), $\bf{R}_{\bf{t}}$, $\bf{Z}_{\bf{t}}$, $\bf{\tilde H}_{\bf{t}}$, $\bf{H}_{\bf{t}}$ donate reset, update, candidate activation and hidden states of the GRU at time step $t$, respectively, $\bf{W_\star}$, $\bf{U_\star}$ donate corresponding weights of $\bf{X_t}$ and $\bf{H_{t-1}}$, $\bf{b_\star}$ donate biases, $\sigma(\bullet)$ and $\varphi(\bullet)$ donate nonlinear activation functions, and in this paper, $\sigma(\bullet)$ is a sigmoid function ($\sigma (x) = \frac{{{1}}}{{1 + {e^{ - x}}}}$), and $\varphi(\bullet)$ is the tangent activation function ($\varphi(x)=\tanh (x) = \frac{{{e^{x}-e^{ - x}}}}{{e^{x} + {e^{ - x}}}}=2\sigma(2x)-1$), ``$\odot$'' donates component-wise multiplication and ``$\cdot$'' donates matrix multiplication.
\par Eq. (\ref{1}) tells us when $\bf{R}_{\bf{t}}$ is close to 0, the previous hidden state $\bf{H_{t-1}}$ would be abandoned and we only use $\bf{X}_{\bf{t}}$ to compute the hidden state $\bf{H_{t}}$. So the reset gate controls whether ignoring the previous hidden state. From Eq. (\ref{hidden}), we can know that when $\bf{Z}_{\bf{t}}$ is close to 1, only new candidate activation $\bf{\tilde H}_{\bf{t}}$ would be used to update the hidden state; and on the contrary, when $\bf{Z}_{\bf{t}}$ approaches to 0, the hidden state $\bf{H_{t}}$ would equal to the previous hidden state $\bf{H_{t-1}}$. So the update gate decides how much the unit updates its hidden state. Becuse of the existence of these two gates, GRU can capture long-term memory as the LSTM do.

\par From the aforementioned contents, we can know that the GRU structure is very similar to LSTM but having several differences: the GRU uses neither a separate memory cell, nor peephole connections and output activations. However, the GRU couples the input and forget gates into an update gate. Moreover, its reset gate only gates the recurrent connections to the input block. When computing the new candidate activation, the GRU controls the information flow from the previous activation. Though much simpler than LSTM, due to the existence of gates, GRU can also effectively capture long-term temporal dependencies and prohibit vanishing or exploding gradients, which makes GRU can be easily trained. Furthermore, we can also stack GRU blocks to increase the depths of the GRU, using the hidden state $h^{(l-1)}_{t}$ of the GRU layer $l-1$ as the input to the GRU layer $l$ (see Fig. \ref{2 layer_fig}).

\subsection{Image Representation}\label{Image Representation}
In Section-\ref{Related Work}, we have known that CNN is very good at image representation because image features extracted by deep CNN are provided with deep semantic information.
In our model, we use the VGG-Net model pre-trained on ImageNet as the image encoder (you can also fine-tune the CNN model using the training dataset if you want). The representation of each image is as follows:
\begin{equation}\label{image representation}
  v = [CN{N_{{\theta _c}}}(I)]\cdot{\bf{W}_I} + {b_I},
\end{equation}
where $I$ donates the image I, $CN{N_{{\theta _c}}}(I)$ is a 4096-dimension feature vector of image I and this feature vector is the output of the last FC-4096 layer (immediately before the classifier which contains a FC-1000 layer and a softmax layer) of the CNN model, $\theta_c$ is the CNN parameters set which is pre-trained on ImageNet and can be fine-tuned with the experimental datasets. The matrix $\bf{W}_I$ has dimensions $4096\times h$ where $h$ is the dimension of the multi-modal embedding space (in fact, it is also the number of the hidden layer's neurons). Because the image representation and sentence representation will be embedded into the same space that is referred to as multi-modal embedding space, the projection matrix  $\bf{W}_I$ should transform the image feature vector into $h$ dimensions. $b_{I}$ donates the bias.

\subsection{Sentence Representation}\label{Sentence Representation}
Given an image $I$ and its true sentence description $S=(w_0, w_1,\cdots ,w_N)$, the sentence representation is represented by word vector and embedding matrix of sentences. The concrete formula is as follows:
\begin{equation}\label{sentence representation}
  s_t = w_t\cdot{\bf{W}_s},\ t\in \{0,1,\cdots ,N-1\},
\end{equation}
where $w_t$ ($t=0,1,\cdots ,N-1$) donate the word vector of the $t$-th word whose dimension is the size of the dictionary. If we assume the dictionary size equals to $N_0$ (in other words, the training set has $N_0$ different words), $w_t$ is a one-hot $N_0$-dimension vector whose $t$-th element is equal to 1 and others are equal to 0. $w_0$ and $w_N$ donate a special start word and stop word respectively which are the start and end tokens of the sentence. $\bf{W}_s$ is the embedding matrix of sentences which projects the word vector into the embedding space. So the projection matrix $\bf{W}_s$ is a $N_0\times h$ matrix where $N_0$ is the size of the dictionary and $h$ is the dimension of the embedding space.

\subsection{Multi-Modal GRU for Generating Descriptions}\label{m-GRU}
In previous work, language models based on RNNs have shown powerful capabilities to generate a target sentence. These models define a probability distribution $P(T|S)$ in which $T$ donates the target sentence and $S$ donates source sentence and every time only one target word is generated. In our image description task, we also propose a probabilistic framework to generate description conditioning on an image by extending the form of probability distribution (i.e. $P(S|I)$). We compute this probability by the chain rule and the formula is as follows:
\begin{equation}\label{chain rule}
P\left( {S\left| I;\theta \right.} \right) = \prod\limits_{t = 0}^N {P\left( {\left. {{w_t}} \right|I,{w_0}, \cdots ,{w_{t - 1}};\theta} \right)},
\end{equation}
where $\theta$ represents the parameters of our model, including aforementioned $\bf{W}_I$, $\bf{W}_s$, $b_I$ and all the weights and biases of GRU. At the training step, $(S,I)$ is a training pair and the parameter set $\theta$ is trained through maximizing $P(S|I)$  over the whole training set.
\par In our model, $P(w_t|I, w_0, \cdots, w_{t-1})$ is modeled by the GRU block, the concrete formula is as follows:
\begin{equation}\label{x}
{x_t} = \left\{ {\begin{array}{*{20}{l}}
{v,\ \ \ \ {\rm{if}}\ \  t =  - 1}\\
{{s_t},\ \ \ {\rm{otherwise}}\ \ t \in \left\{ {0,1, \cdots ,N - 1} \right\} }
\end{array}} \right.,
\end{equation}
\begin{equation}\label{h}
{h_t} = {\rm{GRU}}({x_t},{h_{t - 1}}),\ \ t \in \left\{ {0,1, \cdots ,N - 1} \right\},
\end{equation}
\begin{equation}\label{y}
{y_{t + 1}} = {h_t} \cdot {{\bf{W}}_{\bf{d}}} + {b_d},\ \ t \in \left\{ {0,1, \cdots ,N - 1} \right\},
\end{equation}
\begin{equation}\label{p}
{p_{t + 1}} = {\rm{softmax(}}{y_{t + 1}}),\ \ t \in \left\{ {0,1, \cdots ,N - 1} \right\},
\end{equation}
where $p_{t+1}$ is representation  for $P(w_{t+1}|I, w_0, \cdots, w_t)$. Eqs. (\ref{y}) and (\ref{p}) show that the GRU state $h_t$ is fed into a softmax layer which will produce a probability distribution over all words. Projection matrix $\bf{W_d}$ has dimensions $h\times N_0$, so $p_{t+1}$ is a $N_0$ dimension vector whose each element donates the word probability.
\par We map the image and the sentence into the same space by using a vision deep CNN and word embedding matrix $\bf{W}_s$ (see Section-\ref{Image Representation} and Section-\ref{Sentence Representation}). The image $I$ input to the GRU network only at the time step $t=-1$. Vinyals \textsl{et al.} \cite{vinyals2015nic} and Karparthy \textsl{et al.} \cite{karpathy2015devs} have found that the image $I$ fed into the multi-modal only once is much better than at each time step. That's because the noise and overfitting problem would be introduced into the designed model. Our cost function is settled as the negative likelihood of the correct word at each time step and its formula is as follows:
\begin{equation}\label{loss}
C(S,I;\theta ) =  - \sum\limits_{t = 1}^N {\log {p_t}} + \lambda_{\theta}\cdot\left\| \theta  \right\|_{2}^2.
\end{equation}
Note that $\lambda_{\theta}\cdot\left\| \theta  \right\|_{2}^2$ is a regularization term.
\par Through minimizing the cost function $C(\theta)$, all the parameters of our model will be optimized. At the training step, we compute $h_{-1}=GRU(x_{-1})$ and $h_0 = GRU(x_0,h_{-1})$. In other words, $h_{-1}$ only includes the image contents, $x_0$ is the start vector and we compute the distribution over the first word $p_1$. By the same token, we compute every word distribution and maximize the condition probability described in Eq. (\ref{chain rule}). At the testing time, to generate a description, we compute $h_{-1}=GRU(x_{-1})$, set $x_0$ as the start vector and compute the distribution $p_1$. We select the argmax as the first word $w_1$ and then set its representation as $x_2$, then predict the second word $w_2$. This process is repeated until the end token $w_N$ is generated.

\section{Experiments} \label{Experiments}

In this section, the effectiveness of our model on image description is evaluated by several metrics. We begin by describing the datasets used for training and testing. Then we introduce some standard metric methods and compare our model with other models using theses metrics.
\subsection{Datasets}\label{Dataset}
In this section, three benchmark datasets are introduced. They are Flickr8K \cite{rashtchian2010flickr8k}, Flickr30K \cite{young2014flickr30k} and MS COCO \cite{lin2014coco}.
\begin{figure*}[htbp]
  \centering
  \includegraphics[width=0.85\linewidth]{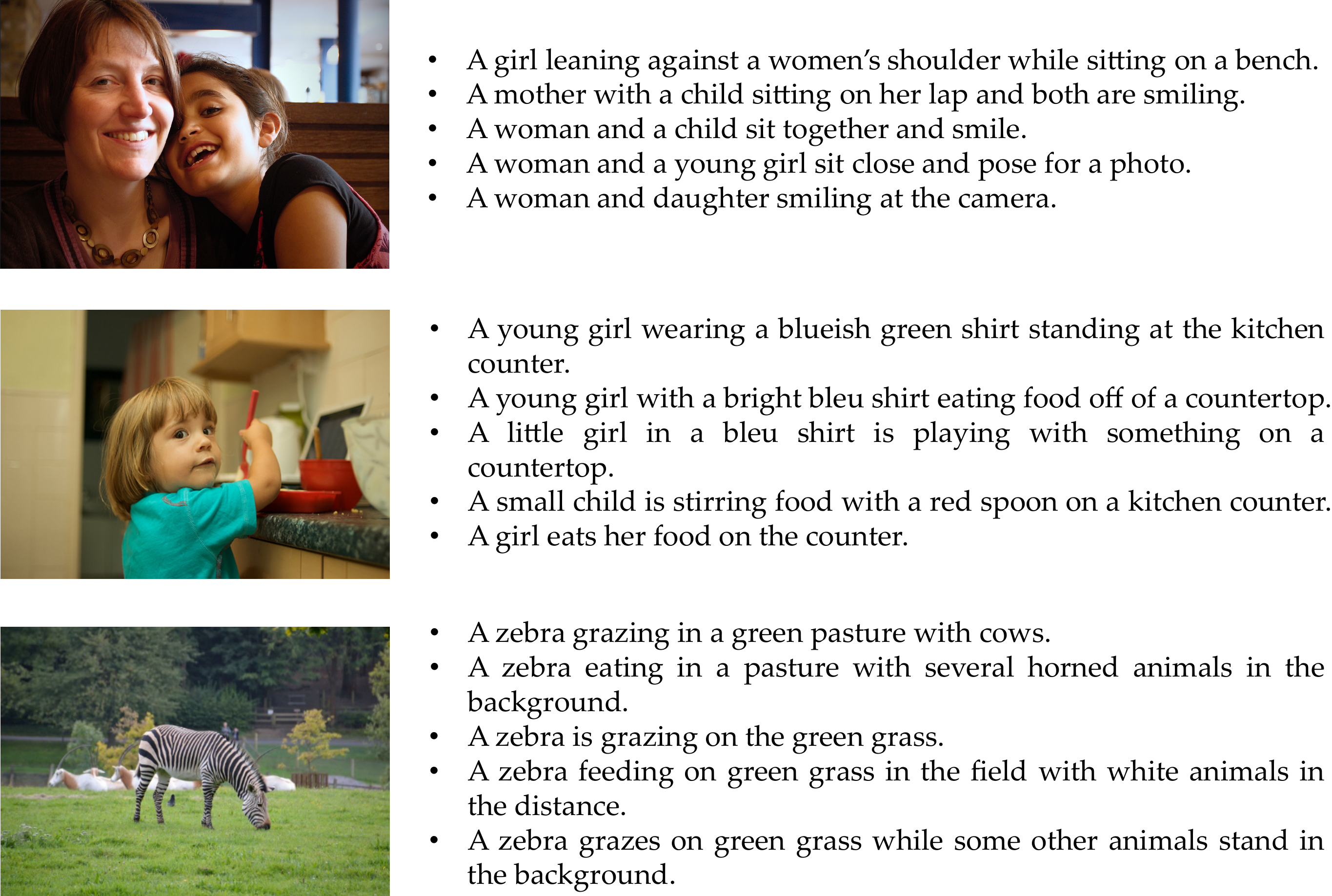}
  \caption{Examples of image-description pairs of the three benchmark datasets. From the top to the bottom, they are come from the Flickr8K dataset, the Flickr30K dataset and the MS COCO dataset. Each image from these three datasets has five natural language sentences to describe the content of the image.}\label{fig.2}
\end{figure*}

\begin{itemize}
\item \textbf{Flickr8K.} The Flickr8K dataset consists of 8,092 images obtained from the ``flickr'' websites. The images in this dataset focus on people and animals (mainly dogs) performing some actions. Each image in the Flickr8K has five sentence annotations generated by \textsl{Amazon Mechanical Turk} (AMT). Every sentence should contain objects, scenes, their attributions and activities shown in the image. An example image-description pair is shown in the top row of Fig. \ref{fig.2}. The Flickr8K dataset is split into three disjoint sets: the training set consists of 6,000 image-description pairs, the validation set consists of 1,000 image-description paris and testing set consists of 1,000 image-description pairs. In our experiments, we adopt the standard separation.
\item \textbf{Flickr30K.} This dataset is a extension of Flickr8K, so the grammar and style for the describing sentences of the dataset are similar to Flickr8K. It consists of 31,783 images and 158,915 describing sentences that focus on everyday activities, events and scenes. An example image-description pair is shown in the middle row of Fig. \ref{fig.2}. We adopt the separation of training, validation and testing sets following the previous work \cite{karpathy2015devs} , \textsl{i.e.}, 1,000 image-description pairs for both testing and validation, and others for training.
\item \textbf{MS COCO.} This dataset is continuously updated and can be used in many different tasks such as image description and image retrieval. The vocabulary in MS COCO is much bigger, more different and has a larger mismatch than Flickr30K. Recently, MS COCO becomes the biggest and highest quality dataset for image description. An example image-description pair can be found in the bottom row of Fig. \ref{fig.2}. The dataset consists of 82,783 training images, 40,504 validation images and 40775 testing images. Each image in training and validation sets has 5 correlation sentence descriptions, but the testing images have no sentence description. Because the test annotations are not available, we randomly sampled 5,000 images from all the validation set for both validation and testing.
\end{itemize}
\par Before the experiment, we have preprocessed the datasets as \cite{karpathy2015devs} did. At first, we convert all letters of sentences to lowercase and remove non-alphanumeric characters. Then we get rid of words that occur less than five times on the training set.
\subsection{Evaluation Metrics}
We evaluate our model on both bidirectional image and sentence retrieval and image description generation. Before we show the results of experiments, we will introduce some metrics for bidirectional image-sentence retrieval and evaluating the quality of the generated sentence.
We use the R@K and Med$r$ \cite{hodosh2013framing} metrics for retrieval and BLEU \cite{papineni2002bleu}, METEOR \cite{banerjee2005meteor} and CIDEr \cite{vedantam2015cider} to evaluate the quality of sentence generated by our model.
\begin{itemize}
\item \textbf{R@K.} \textsl{Recall at position $k$} (R@K) is the recall rate of a correctly retrieved groundtruth given top K candidates. It measures the fraction of times a correct item was found among the top K results. Because for each query, the gold item is either found among the top K results or not, R@K is a binary metric. To compare with results of other models, we set $K\in\left\{ 1,5,10 \right\}$ (\textsl{i.e.} R@1, R@5, R@10). A higher R@K is, a better retrieval performance is. Given $N$ images (or sentences), according to the corresponding retrieved sentences (or images), R@K is defined as follows:
\begin{equation}\label{R@K}
 R@K = \frac{1}{N}\sum\limits_{i = 1}^N {\mathds{1}({r_i} \le K)},
\end{equation}
where $r_i$ is the number of the correct result for the query $i$, $\mathds{1}(\bullet)$ donates the indicate function and it is defined as follows:
\[\mathds{1}({r_i} \le K) = \left\{ {\begin{array}{*{20}{l}}
{1,\ \ {\rm{if}}\ \left\{ {{r_i}\mid{r_i} \le K} \right\} \ne \emptyset }\\
{0,\ \ \rm{otherwise}}
\end{array}.} \right.\]
\item \textbf{Med$r$.} \textsl{Median rank} (Med$r$) is another metric for retrieval. Contrary to R@K, the Med$r$ donates the K at which a system has a recall of 50\% and evaluates the overall performance on retrieval. Med$r$ is defined as follows:
\begin{equation}\label{Med$r$}
 {\rm{Med}}r = \frac{1}{N}\sum\limits_{i = 1}^N {r_i^{med}},
\end{equation}
where $N$ is the number of queries and $r_i^{med}$ is the value of the median ranked correct result of query $i$. A lower Med$r$ indicates a better performance.
\item \textbf{BLEU.} BLEU is short for BiLingual Evaluation Understudy. This metric is one of the first metrics to achieve a high correlation with human judgements of quality. BLEU scores (\textsl{i.e.} B-1, B-2, B-3 and B-4) are widely used in MT and it is a modified form of precision to compare  N-gram (up to 4-gram) fragments of the hypothesis translation with a set of good quality reference translation. We consider image description task as a ``translation'' problem, so we use BLEU as one of metrics to evaluate descriptions generated by our model. To calculate BLEU scores, we should first compute the modifier n-gram precision $p_n$. And then we compute the geometric mean of $p_n$ up to $N$ ($N=1,2,3,4$) and multiply a penalty $BP$:
\begin{equation}\label{B-N penalty}
  BP =  - \min (1,{e^{1 - \frac{r}{c}}}),
\end{equation}
where $r$ and $c$ donate the length of reference and generated sentence, respectively. Then the B-N is defined as follows:
\begin{equation}\label{B-N}
  B - N = BP\cdot\exp \left\{ {\frac{1}{N}\sum\limits_{n = 1}^N {\log {P_n}} } \right\}.
\end{equation}

\par Through Eq. (\ref{B-N}) we know that B-1 accounts for information retained by the generated description, while B-2, B-3 and B-4 account for the fluency of the sentence. And much higher a BLUE score donates our description for images is much better.
\item \textbf{METEOR.} METEOR is short for Metric for Evaluation of Translation with Explicit ORdering. This metric is based on the harmonic mean of unigram precision and recall, with recall weighted higher than precision, which is very different from BLEU that only considers precision. Another difference is METEOR seeks correlation with human judgement at the sentence or segment level, while BLEU seeks correlation at the corpus level. Moreover, METEOR extends word matches which includes similar words based on WordNet synonyms and stemmed tokens. When we calculate METEOR score, we need three steps: at first, we should calculate the precision $P$ and recall $R$, and then calculate their harmonic mean $H$; secondly, we calculate the penalty $PM$; at last, the METEOR score $M$ will be computed. The formula is as follows:
\begin{equation}\label{H}
  H = \frac{{10PR}}{{R + 9P}},
\end{equation}
\begin{equation}\label{PM}
  PM = 0.5*{\left( {\frac{C}{{Um}}} \right)^3},
\end{equation}
\begin{equation}\label{M}
  M = H*(1 - PM),
\end{equation}
where $C$ is the number of chunks and $Um$ is the number of matched unigrams (If you want to know more information of these parameters, you can see the reference \cite{banerjee2005meteor}). As same as the BLEU score, a higher METEOR score is better.
\item \textbf{CIDEr.} This metric is a very new metric for image description and CIDEr is short for Consensus-based Image Description Evaluation. The metric can inherently capture sentence similarity, the notions of grammaticality, saliency precision and recall. There are three motivations for CIDEr metric: firstly, a measure should encode the frequency n-grams in the hypothesis sentence are in the reference sentences; on the contrary, if n-grams are not present in the reference sentences, they should be present in the candidate sentence; thirdly, the more n-grams commonly occur in the dataset, the less information is included in these n-grams, so they should be given lower weight. So the CIDEr is more intricate but more reasonable than BLEU and METEOR metrics.  To calculate CIDEr score, we should first compute the \textsl{Term Frequency Inverse Document Frequency} (TF-IDF) weighting. Then CIDEr score is computed using the cosine similarity between the hypothesis sentence and reference sentences. The detail formula can be seen from the reference \cite{vedantam2015cider}.
\end{itemize}

\subsection{Comparison Models}\label{comparison model}
In this section, some typical models which correlate to image description are briefly introduced. These  models are chosen as comparative approaches because they represent different methods for image description task.

\begin{itemize}
\item \textbf{DT-RNN.} The \textsl{dependency tree recursive neural network} (DT-RNN) is proposed by R. Socher \textsl{et al.} \cite{socher2014dt-rnn}. DT-RNN model is based on constituency tree. More particularly, this model exploits dependency trees to embed sentences into a vector space. Firstly, the sentence representation is learnt based on its grammatical dependency tree and the image representation is extracted by deep CNN. Then these two vectors are mapped into the same semantic space and evaluating the similarity between the image and sentence by measuring the distance in that space. The DT-RNN has a fixed language template, so sentences generated by this model may have less variety.
\item \textbf{DeViSE.} The \textsl{deep visual-semantic embedding} (DeViSE) model \cite{frome2013devise} leverages textual data to learn semantic relationships between labels and images. It is a word-level inter-modal correspondences between image and semantic. DeViSE treat each word equally and averages ($L_2$ normalized) their word vectors as the representation of the sentence. CNN is used as the image feature detector in the DeViSE model. This model uses the hinge rank loss function as the objective function to evaluate the similarity between images and semantics.
\item \textbf{KCCA.} \textsl{Kernel canonical correlation analysis} (KCCA) is an extension of \textsl{canonical correlation analysis} (CCA) \cite{andrew2013deep} which finds linear correlation between the data pairs. CCA uses linear projection which may not capture the intrinsic structures while are necessary to explain the correlation of multi-modal data. KCCA uses kernel functions that map the original items into high-order spaces. Though KCCA has achieved success to associate images with individual words or annotations \cite{socher2014dt-rnn}, a fatal flaw is that it requires two kernel matrices of training data to be stored in the memory --this is prohibitive if the dataset is very large.
\item \textbf{DeFrag.} \textsl{Deep fragment} embedding \cite{karpathy2014deep} exploits R-CNN model detector to detect objects in all images. Descriptions are presentented as dependency trees by the Stanford CoreNLP parser. DeFrag is a phrase-level inter-modal correspondences between image and text, it learns the correlation of textual entities and visual objects, so it neglects the compositional semantics in the pairs of image and descriptions.
\item \textbf{m-RNN.} The \textsl{multi-modal recurrent neural network} (m-RNN) \cite{mao2014m-rnn} consists of two sub-networks: a deep CNN for images and a deep RNN for descriptions, which is similar to our model. This model uses perplexity to bridge  images and descriptions. The m-RNN dose not use a ranking loss which is used in aforementioned models but instead with log-likelihood loss which which predicts the net word in a sequence conditioned on image. This implicit only a word is generated in one time. The m-RNN model gets a lot of breakthroughs in image description task, but the RNN they used is not so strong in long-term memory.
\item \textbf{MNLM.} The \textsl{multi-modal neural language model} (MNLM) \cite{kiros2014mnlm} is a encoder-decoder model which learns a joint image-sentence embedding where sentences using LSTM recurrent neural network. Image features extracted by deep CNN are projected into the embedding space of LSTM hidden states. The \textsl{structure-content neural language model } (SC-NLM) is the decoder which generates novel descriptions from scratch. LSTM can solve the long-term memory problem through the gate technology, but every gate may bring one time parameters more than the villa RNN model, this may lead to overfitting problem or at least take more time to train the model.
\item \textbf{LRCN.} The \textsl{long-term recurrent convolutional network} (LRCN) is proposed by J. Donahue \textsl{et al.} \cite{donahue2015lrcn}. This model is a recurrent convolutional architecture suitable for large-scale visual learning and it can be used in video recognition tasks, image description and retrieval problems. They use two layers LSTM RNN model and image features extracted by a deep CNN are input into the sequential model at each timestep, which are different with our model. Our model input image feature into sequential model only at the first timestep. LRCN gains a lot of success, but the structure is complex and the number of parameters is so huge that makes it need more time to train.
\end{itemize}

\begin{figure*}
  \centering
  \includegraphics[width=0.55\linewidth]{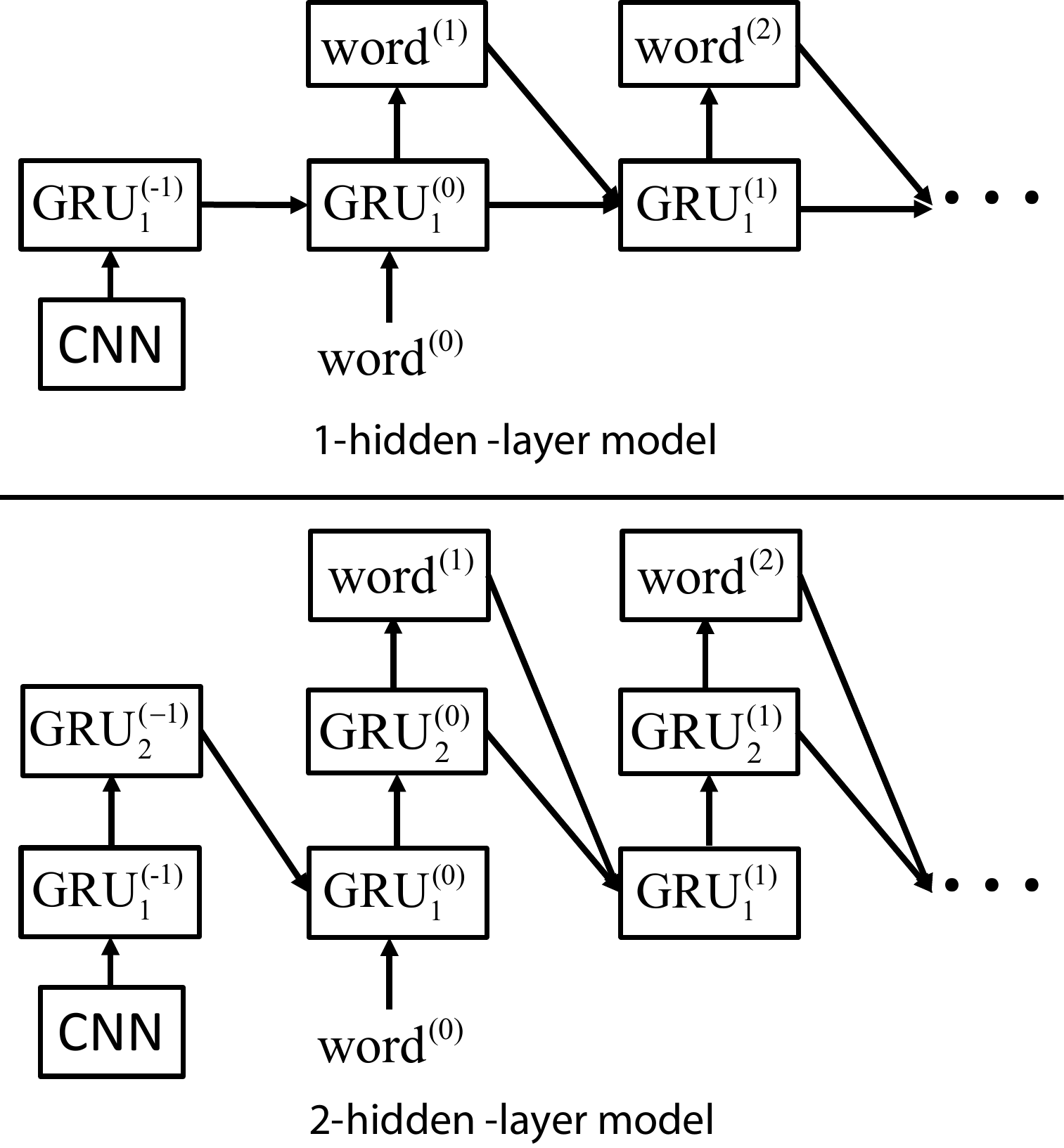}\\
  \caption{Two variations of our multi-modal GRU architecture. The top one has one hidden layer and the bottom one has two hidden layers, so we call them 1-layer model and 2-layer model, respectively. We compare these two variations on the Flickr30K dataset.}
  \label{2 layer_fig}
\end{figure*}

\subsection{Experimental Results} \label{eperimental results}

In this section, we show the bidirectional retrieval results and generation results compared with the aforementioned state-of-the-art models. We have two architectural variants showed in Fig. \ref{2 layer_fig} to research how different GRU layers impact the results of image description. We name the two variants as 1-layer model (the top one) and 2-layer model (the bottom one), respectively. As illustrated in \cite{donahue2015lrcn}, increasing the number of the RNN layer can improve the performance of RNN because the multi-layer RNN can explore the nonlinear relationship between different modal. Furthermore, compared with conventional stacking RNN, the number of parameters in our 2-layer model is half less than the 2-layer model with traditional stacking strategy. In other words, our model can attain higher performance with less parameter. In our experiments, we only train the 2-layer model on the Flickr30K dataset but not on the Flickr8K and MS COCO datasets, because the MS COCO dataset need too much time to train and data in the Flickr8K dataset is a little few which easily leads to overfitting.

\par Some researchers also use stacked-RNN to explore more complexity nonlinearity between different modal data, but we have an important variant. Fig. \ref{2 layer_fig} shows the difference between conventional stacked RNN and our stacked RNN. Information of the both models flows from the previous step to next step and the lower hidden state to upper hidden state. For conventional stacked RNN, however, the previous step upper hidden state does not be imported into the next lower hidden state. The hidden state of conventional stacked RNN is computed by

\begin{equation}\label{s-RNN}
h_j^{(t)} = \varphi ({{\bf{W}}_{j - 1 \to j}}h_{j - 1}^{(t)} + {{\bf{U}}_{j \to j}}h_j^{(t - 1)}), (j = 1, 2),
\end{equation}

where subscript $_{i \to j}$ donates the transition from layer $i$ to alyer $j$. When $j=1$, $h_{j - 1}^t= x_t$. But our stacked method donates this format:

\begin{equation}\label{our-RNN}
\left\{ {\begin{array}{*{20}{l}}
{h_1^{(t)} = \varphi ({{\bf{W}}_{0 \to 1}}{x_t} + {{\bf{U}}_{2 \to 1}}h_2^{(t - 1)})}\\
{h_2^{(t)} = \varphi ({{\bf{W}}_{1 \to 2}}h_1^{(t)})}
\end{array}} \right..
\end{equation}

From Eq. (\ref{our-RNN}), we know that the previous upper layer hidden state is considers as one of the input of the next step hidden state. So there is a feedback in our stacked RNN. When the number of hidden units per layer, the conventional stacked RNN has more parameters than ours.

\subsubsection{Image-Sentence Retrieval Results}

\begin{table*}[!htb] \normalsize \addtolength{\tabcolsep}{-3.5pt}\setlength{\tabcolsep}{5.5pt}
\setlength{\abovecaptionskip}{10pt}
\setlength{\belowcaptionskip}{5pt}
\begin{spacing}{1.5}
\newcommand{\tabincell}[2]{\begin{tabular}{@{}#1@{}}#2\end{tabular}}
\centering
\caption{Bidirectional image and sentence retrieval results on Flickr8K \& MS COCO}
\label{Retrieval Results Flickr8K}
\small\begin{tabular}{l|cccc|cccc}
  \hline
 &\multicolumn{4}{c|}{Sentence Retrieval}  &\multicolumn{4}{c}{Image Retrieval}\\
  \cline{2-9}\raisebox{0.5em}{Model} & \tabincell{c}{R@1} & \tabincell{c}{R@5} & \tabincell{c}{R@10} & \tabincell{c}{Med$r$} &\tabincell{c}{R@1} & \tabincell{c}{R@5} & \tabincell{c}{R@10} & \tabincell{c}{Med$r$}\\
  \hline
  \multicolumn{9}{c}{\textbf{Flickr8K}}\\
  \hline
  \raisebox{0em} {DT-RNN \cite{socher2014dt-rnn}} & 4.5 & 18.0 & 28.6 & 32 & 6.1 & 18.5 & 29.0 & 29\\
  \raisebox{0em} {DeViSE \cite{frome2013devise}} & 4.8 & 16.5 & 27.3 & 28 & 5.9 & 20.1 & 29.6 & 29\\
  \raisebox{0em} {KCCA \cite{socher2014dt-rnn}} & 8.3 & 21.6 & 30.3 & 34 & 7.6 & 20.7 & 30.1 & 38\\
  \raisebox{0em} {DeFrag \cite{karpathy2014deep}} & 5.9 & 19.2 & 27.3 & 34 & 5.2 & 17.6 & 26.5 & 32\\
  \raisebox{0em} {m-RNN \cite{mao2014m-rnn}} & 14.5 & 37.2 & 48.5 & 11 & 11.5 & 31.0 & 42.4 & 15\\
  \raisebox{0em} {MNLM \cite{kiros2014mnlm}} & 13.5 & 36.2 & 45.7 & 13 & 10.4 & 31.0 & 43.7 & 14\\
  \raisebox{0em} {MNLM-VGG \cite{kiros2014mnlm}} & 18.0 & 40.9 & 55.0 & 8 & 12.5 & 37.0 & 51.5 & 10\\
  \raisebox{0em} {DeVS-DepTree \cite{karpathy2015devs}} & 14.8 & 37.9 & 50.0 & 9 & 11.6 & 31.4 & 43.8 & 13\\
  \raisebox{0em} {DeVS-BRNN \cite{karpathy2015devs}} & 16.5 & 40.6 & 54.2 & 8 & 11.8 & 32.1 & 44.7 & 12\\
  \raisebox{0em} {Google-NIC \cite{vinyals2015nic}} & 20.0 & - & 61.0 & \textbf{6} & 19.0 & - & 64.0 & \textbf{5}\\
  \hline
   \raisebox{0em} {Ours: 1-layer} & \textbf{24.3} & \textbf{47.7} & \textbf{64.2} & \textbf{6} & \textbf{20.8} & \textbf{35.5} & \textbf{66.1} & \textbf{5}\\
  \hline
  \multicolumn{9}{c}{\textbf{MS COCO}}\\
   \hline
   \raisebox{0em} {DeViSE \cite{frome2013devise}} & 2.1 & 10.6 & 17.3 & 61 & 2.7 & 11.5 & 19.4 & 45\\
   \raisebox{0em} {DeFrag \cite{karpathy2014deep}} & 25.8 & 56.3 & 70.8 & 4 & 19.8 & 48.4 & 63.8 & 6\\
   \raisebox{0em} {DeVS-BRNN \cite{karpathy2015devs}} & 16.5 & 39.2 & 52.0 & 9 & 10.7 & 29.6 & 42.2 & 14\\
  \raisebox{0em} {m-RNN \cite{mao2014m-rnn}} & 41.0 & \textbf{63.8} & 73.7 & 3 & 22.8 & 50.7 & 63.1 & 5\\
  \hline
  \raisebox{0em} {Ours: 1-layer} & \textbf{42.7} & 63.1 & \textbf{78.1} & \textbf{3} & \textbf{30.8} & \textbf{53.2} & \textbf{64.9} & \textbf{4}\\
  \hline
\end{tabular}
\end{spacing}
\end{table*}

\begin{table*}[!htb] \normalsize \addtolength{\tabcolsep}{-3.5pt}\setlength{\tabcolsep}{5.5pt}
\setlength{\abovecaptionskip}{10pt}
\setlength{\belowcaptionskip}{5pt}
\begin{spacing}{1.5}
\newcommand{\tabincell}[2]{\begin{tabular}{@{}#1@{}}#2\end{tabular}}
\centering
\caption{Bidirectional image and sentence retrieval results on Flickr30K}
\label{Retrieval Results Flickr30K}
\small\begin{tabular}{l|cccc|cccc}
  \hline
  \multicolumn{9}{c}{\textbf{Flickr30K}}\\
  \hline
 &\multicolumn{4}{c|}{Sentence Retrieval}  &\multicolumn{4}{c}{Image Retrieval}\\
  \cline{2-9}\raisebox{0.5em}{Model} & \tabincell{c}{R@1} & \tabincell{c}{R@5} & \tabincell{c}{R@10} & \tabincell{c}{Med$r$} &\tabincell{c}{R@1} & \tabincell{c}{R@5} & \tabincell{c}{R@10} & \tabincell{c}{Med$r$}\\
  \hline
  \raisebox{0em} {DT-RNN \cite{socher2014dt-rnn}} & 9.6 & 29.8 & 41.1 & 16 & 8.9 & 29.8 & 41.1 & 16\\
  \raisebox{0em} {DeViSE \cite{frome2013devise}} & 4.5 & 18.1 & 29.2 & 26 & 6.7 & 21.9 & 32.7 & 25\\
  \raisebox{0em} {DeFrag \cite{karpathy2014deep}} & 14.2 & 37.7 & 51.3 & 10 & 10.2 & 30.8 & 44.2 & 14\\
  \raisebox{0em} {m-RNN \cite{mao2014m-rnn}} & 18.4 & 40.2 & 50.9 & 10 & 12.6 & 31.2 & 41.5 & 16\\
  \raisebox{0em} {MNLM \cite{kiros2014mnlm}} & 14.8 & 39.2 & 50.9 & 10 & 11.8 & 34.0 & 46.3 & 13\\
  \raisebox{0em} {MNLM-VGG \cite{kiros2014mnlm}} & 23.0 & 50.7 & 62.9 & 5 & 16.8 & 42.0 & 56.5 & 8\\
  \raisebox{0em} {DeVS-DepTree \cite{karpathy2015devs}} & 20.0 & 46.6 & 59.4 & 5 & 15.0 & 36.5 & 48.2 & 10\\
  \raisebox{0em} {DeVS-BRNN \cite{karpathy2015devs}} & 22.2 & 48.2 & 61.4 & 5 & 15.2 & 37.7 & 50.5 & 9\\
  \raisebox{0em} {LRCN \cite{donahue2015lrcn}} & 14.0 & 34.9 & 47.0 & 11 & 17.5 & 40.3 & 50.8 & 9\\
  \raisebox{0em} {Google-NIC \cite{vinyals2015nic}} & 17.0 & - & 56.0 & 7 & 17.0 & - & 57.0 & 7\\
  \hline
  \raisebox{0em} {Ours: 1-layer} & 22.9 & 51.2 & 60.0 & 7 & 16.1 & 40.7 & 50.0 & 7\\
  \raisebox{0em} {Ours: 2-layer} & \textbf{25.4} & \textbf{51.9} & \textbf{65.2} & \textbf{4} & \textbf{19.4} & \textbf{59.8} & \textbf{57.0} & \textbf{6}\\
  \hline

\end{tabular}
\end{spacing}
\end{table*}

Ranking experiments are not very suitable for evaluating description for images, but many correlating paper reports ranking scores, so  we also use the testing sentences as candidates to rank the test image. Table \ref{Retrieval Results Flickr8K} shows the bidirectional image and sentence retrieval results on the Flickr8K dataset and the MS COCO dataset. Except our 1-layer model, the Google-NIC model \cite{vinyals2015nic} performs best on these two datasets. However, our 1-layer model gets better results on these ranking metrics than the Google-NIC model (only a few of score ranked the second). Considering that the Google-NIC using LSTM as the multi-modal embedding network and the RNN dimension is 512 (\textsl{i.e.} each hidden layer has 512 neurons) which is equal to ours, this model for MS COCO has 1,000,000 more parameters than ours. So they need more time to train the model. Table \ref{Retrieval Results Flickr30K} lists the bidirectional image and sentence retrieval results on the Flickr30K dataset. Although the MNLM-VGG model gets better score in several metrics, our 1-layer model also shows good performance on this dataset. Moreover, our 2-layer model almost refreshes all the scores reported in the state-of-the-art models. Through these two tables, we can know that the ``CNN+RNN'' model (our model, m-RNN \cite{mao2014m-rnn}, DeVS \cite{karpathy2015devs}, Google-NIC \cite{vinyals2015nic} are all this type) shows a strong ability for the image description problem.

\subsubsection{Generation Results}

\begin{table}[!htb] \normalsize \addtolength{\tabcolsep}{-3.5pt}\setlength{\tabcolsep}{2.5pt}
\setlength{\abovecaptionskip}{10pt}
\setlength{\belowcaptionskip}{5pt}
\begin{spacing}{1.5}
\newcommand{\tabincell}[2]{\begin{tabular}{@{}#1@{}}#2\end{tabular}}
\centering
\caption{Results of generated sentences on the Flickr8K \& MS COCO}
\label{Generated Results Flickr8K}
\small\begin{tabular}{l|ccccccc}

  \hline
  \raisebox{0em}{Model} & \tabincell{c}{B-1} & \tabincell{c}{B-2} & \tabincell{c}{B-3} & \tabincell{c}{B-4} &\tabincell{c}{METEOR} & \tabincell{c}{CIDEr}\\
  \hline
  \multicolumn{7}{c}{\textbf{Flickr8K}}\\
  \hline
  \raisebox{0em} {m-RNN \cite{mao2014m-rnn}} & 56.5 & 38.6 & 25.6 & 17.0 & - & - \\
  \raisebox{0em} {DeVS \cite{karpathy2015devs}} & 57.9 & 38.3 & 24.5 & 16.0 & 16.7 & \textbf{31.8} \\
  \raisebox{0em} {LRVR \cite{chen2015lrvr}} & - & - & - & 14.1 & 18.0 & - \\
  \raisebox{0em} {Google-NIC \cite{vinyals2015nic}} & \textbf{63.0} & 41.0 & 27.0 & - & - & - \\
  \hline
  \raisebox{0em} {Ours: 1-layer} & 62.9 & \textbf{41.2} & \textbf{29.2} & \textbf{21.0} & \textbf{18.5} & 30.1 \\
  \hline
  \multicolumn{7}{c}{\textbf{MS COCO}}\\
  \hline
  \raisebox{0em} {m-RNN \cite{mao2014m-rnn}} & 66.8 & \textbf{48.8} & 34.2 & 23.9 & 22.1 & 72.9 \\
  \raisebox{0em} {DeVS \cite{karpathy2015devs}} & 62.5 & 45.0 & 32.1 & 23.0 & 19.5 & 66.0 \\
  \raisebox{0em} {LRVR \cite{chen2015lrvr}} & - & - & - & 19.0 & 20.4 & - \\
  \raisebox{0em} {Google-NIC \cite{vinyals2015nic}} & 66.6 & 46.1 & 32.9 & \textbf{24.6} & 23.7 & \textbf{85.5 }\\
  \raisebox{0em} {LRCN \cite{donahue2015lrcn}} & 62.8 & 44.2 & 30.4 & 21 & - & - \\
  \raisebox{0em} {Att-CNN+LSTM \cite{DBLP:conf/cvpr/WuSLDH16}} & 74 & 56 & 42 & 31 & 26 & 94 \\
  \raisebox{0em} {MAT \cite{DBLP:conf/ijcai/LiuSWWY17}} & 73.1 & 56.7 & 42.9 & 32.3 & 25.8 & 105.8 \\
  \hline
  \raisebox{0em} {LSTM-based} & 66.9 & 48.2 & 32.4 & 22.6 & 24.2 & 81.2 \\
  \hline
  \raisebox{0em} {Ours: 1-layer} & \textbf{67.2} & 47.9 & \textbf{33.1} & 22.6 & \textbf{25.5} & 80.3 \\
  \hline
\end{tabular}
\end{spacing}
\end{table}

\begin{table}[ht] \normalsize \addtolength{\tabcolsep}{-3.5pt}\setlength{\tabcolsep}{3pt}
\setlength{\abovecaptionskip}{10pt}
\setlength{\belowcaptionskip}{5pt}
\begin{spacing}{1.5}
\newcommand{\tabincell}[2]{\begin{tabular}{@{}#1@{}}#2\end{tabular}}
\centering
\caption{The number of parameters in GRU and LSTM}
\label{Generating Results}
\small\begin{tabular}{l|p{1.5cm}<{\centering}|p{1.5cm}<{\centering}|p{1.5cm}<{\centering}}
  \hline
 &\multicolumn{3}{c}{Dimensionality of the Hidden State}\\
  \cline{2-4}\raisebox{0.5em}{} & \tabincell{c}{256} & \tabincell{c}{512} & \tabincell{c}{1000}\\
  \hline
  \raisebox{0em} {GRU} & 393,984 & 1,574,400 & 6,294,528\\
  \raisebox{0em} {LSTM} & 525,312 & 2,099,200 & 8,392,704\\
  \hline
\end{tabular}
\end{spacing}
\end{table}

\begin{table}[!htb] \normalsize \addtolength{\tabcolsep}{-3.5pt}\setlength{\tabcolsep}{2.5pt}
\setlength{\abovecaptionskip}{10pt}
\setlength{\belowcaptionskip}{5pt}
\begin{spacing}{1.5}
\newcommand{\tabincell}[2]{\begin{tabular}{@{}#1@{}}#2\end{tabular}}
\centering
\caption{Results of generated sentences on the Flickr30K dataset}
\label{Generated Results Flickr30K}
\small\begin{tabular}{l|ccccccc}

  \hline
  \multicolumn{7}{c}{\textbf{Flickr30K}}\\
  \hline
  \raisebox{0em}{Model} & \tabincell{c}{B-1} & \tabincell{c}{B-2} & \tabincell{c}{B-3} & \tabincell{c}{B-4} &\tabincell{c}{METEOR} & \tabincell{c}{CIDEr}\\
  \hline
  \raisebox{0em} {m-RNN \cite{mao2014m-rnn}} & 60 & 41 & 28 & 19 & - & - \\
  \raisebox{0em} {DeVS \cite{karpathy2015devs}} & 57.3 & 36.9 & 24.0 & 15.7 & 15.3 & 24.7 \\
  \raisebox{0em} {LRVR \cite{chen2015lrvr}} & - & - & - & 12.6 & 16.4 & - \\
  \raisebox{0em} {LRCN \cite{donahue2015lrcn}} & 58.8 & 39.1 & 25.1 & 16.5 & - & - \\
  \raisebox{0em} {Google-NIC \cite{vinyals2015nic}} & 66.3 & \textbf{42.3} & 27.7 & 18.3 & - & - \\
  \hline
  \raisebox{0em} {Ours: 1-layer} & 60.9 & 38.1 & 25.4 & 16.7 & 25.9 & 43.3 \\
  \raisebox{0em} {Ours: 2-layer} & \textbf{66.9} & 40.5 & \textbf{28.9} & \textbf{20.0} & \textbf{29.5} & \textbf{48.3} \\
  \hline
\end{tabular}
\end{spacing}
\end{table}

Table \ref{Generated Results Flickr8K} shows the results of generated sentences on the Flickr8K dataset and the MS COCO dataset. Results show that our model outperforms all the other models which only used the global image feature without using attention and attribute information on Flickr8K. The Google-NIC model gets a highest score during the comparative models, but our 1-layer model gets higher scores on B-2, B-3, B-4 and METEOR. For the MS COCO dataset, our 1-layer model is comparable to other methods. In fact, the Google-NIC model and the LRCN model's architecture are more complicated than ours, but our model shows a stronger ability than them on the B-1, B-2, B-3 and METEOR metrics. Among all the contrast models, two models show better performance than our model. However, these two models need more extra information and their model are more complicated than ours. First, Att-CNN+LSTM \cite{DBLP:conf/cvpr/WuSLDH16} needs extra attribute information of image. That is to say, it needs to create an attribute dataset to train the attribute predictor. Furthermore, this model should input the attribute vector into the sentence generator, which makes the LSTM needs more parameters. Second, the MAT model need an extra object detector to detect objects in an image. After that, the objects need to be arranged in to a sequence. This stage makes the MAT model is more complicated than us. The performance is also compared with LSTM-based model in Table \ref{Generated Results Flickr8K}. Although the performance is not far away from our 1-layer model, but the number of parameters in our GRU module is a quarter less than that in LSTM \ref{The number of parameters in GRU and LSTM}. So, the LSTM-based model is more calculation amount consuming. Table \ref{Generated Results Flickr30K} shows the results of generated sentences on the Flickr30K dataset. Similar to the aforementioned analysis, our multi-modal GRU model shows a very good performance on the Flickr30K dataset, especially our 2-layer model, which almost archives the best score on all metrics.

\begin{table*}[!htb] \normalsize \addtolength{\tabcolsep}{-3.5pt}\setlength{\tabcolsep}{5.5pt}
\setlength{\abovecaptionskip}{10pt}
\setlength{\belowcaptionskip}{5pt}
\begin{spacing}{1.5}
\newcommand{\tabincell}[2]{\begin{tabular}{@{}#1@{}}#2\end{tabular}}
\centering
\caption{The statistical results on MS COCO}
\resizebox{\linewidth}{!}{
\label{std}
\small\begin{tabular}{|l|c|c|c|c|c|c|c|c|c|c|c|c|}
  \hline
 &\multicolumn{2}{c|}{B-1}  &\multicolumn{2}{c|}{B-2}  &\multicolumn{2}{c|}{B-3}  &\multicolumn{2}{c|}{B-4}  &\multicolumn{2}{c|}{METEOR}  &\multicolumn{2}{c|}{CIDEr}\\
  \cline{2-13}\raisebox{0.5em}{} & \tabincell{c}{Mean} & \tabincell{c}{Std} & \tabincell{c}{Mean} & \tabincell{c}{Std} & \tabincell{c}{Mean} & \tabincell{c}{Std} & \tabincell{c}{Mean} & \tabincell{c}{Std} & \tabincell{c}{Mean} & \tabincell{c}{Std} & \tabincell{c}{Mean} & \tabincell{c}{Std} \\
  \hline
  \hline

  \raisebox{0em} {DeVS} & 60.8 & 1.02 & 42.4 & 1.15 & 30.0 & 0.85 & 21.1 & 1.05 & 18.8 & 0.35 & 57.1 & 1.47 \\
  \raisebox{0em} {Google-NIC} & 64.0 & 0.94 & 45.7 & 0.93 & 32.7 & 0.73 & \textbf{23.9} & \textbf{0.80} & 21.1 & 0.25 & \textbf{76.3} & 1.97 \\
  \hline
  \hline
  \raisebox{0em} {Ours: 1-layer} & \textbf{66.8} & \textbf{0.54} & \textbf{45.9} & \textbf{0.45} & \textbf{33.1} & \textbf{0.63} & 22.9 & 0.85 & \textbf{25.5} & \textbf{0.13} & 73.3 & \textbf{0.98} \\
  \hline

\end{tabular}}
\end{spacing}
\end{table*}

\par To verify the statistical significance of our model, 8 groups experiments on MS COCO have been done to show the statistical characteristics. For each experiment, 5,000 images from original validation set have used to test the models. So all the 40,000 images come from the original validation set have been used. The statistical results have been shown in Table \ref{std}, “Mean” and “Std” denote the mean score and the standard deviation. The results show that our model can get almost the highest indicator and show the most stable performance. Furthermore, it verifies our mode has stronger generalization ability than DeVS and Google-NIC. 

\begin{figure*}[htbp]
  \centering
  \includegraphics[width=0.95\linewidth]{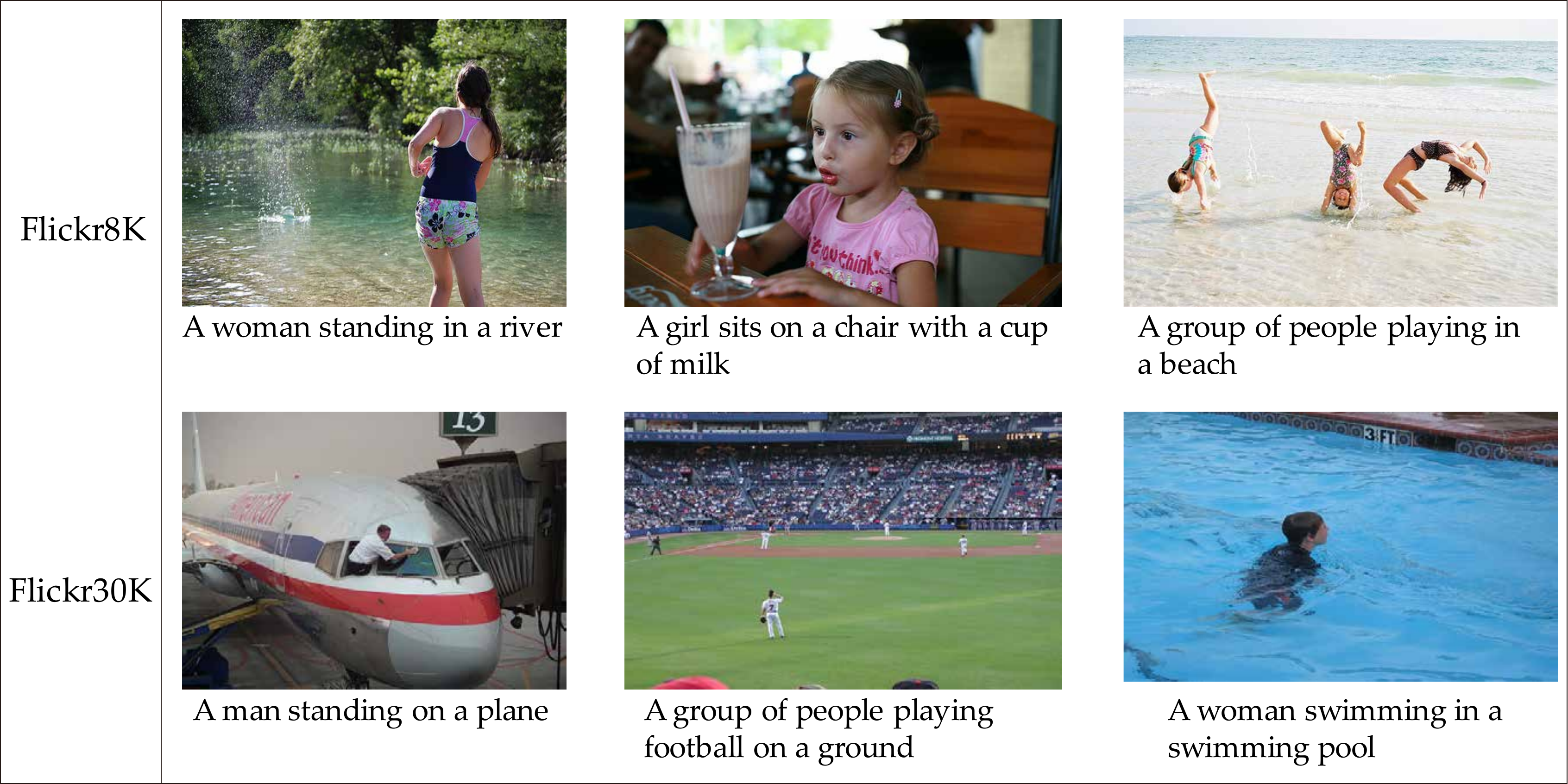}
  \caption{Some visualized captioning results on Flickr8K and Flickr30K.}\label{flickr}
\end{figure*}

\begin{figure*}[htbp]
  \centering
  \includegraphics[width=0.95\linewidth]{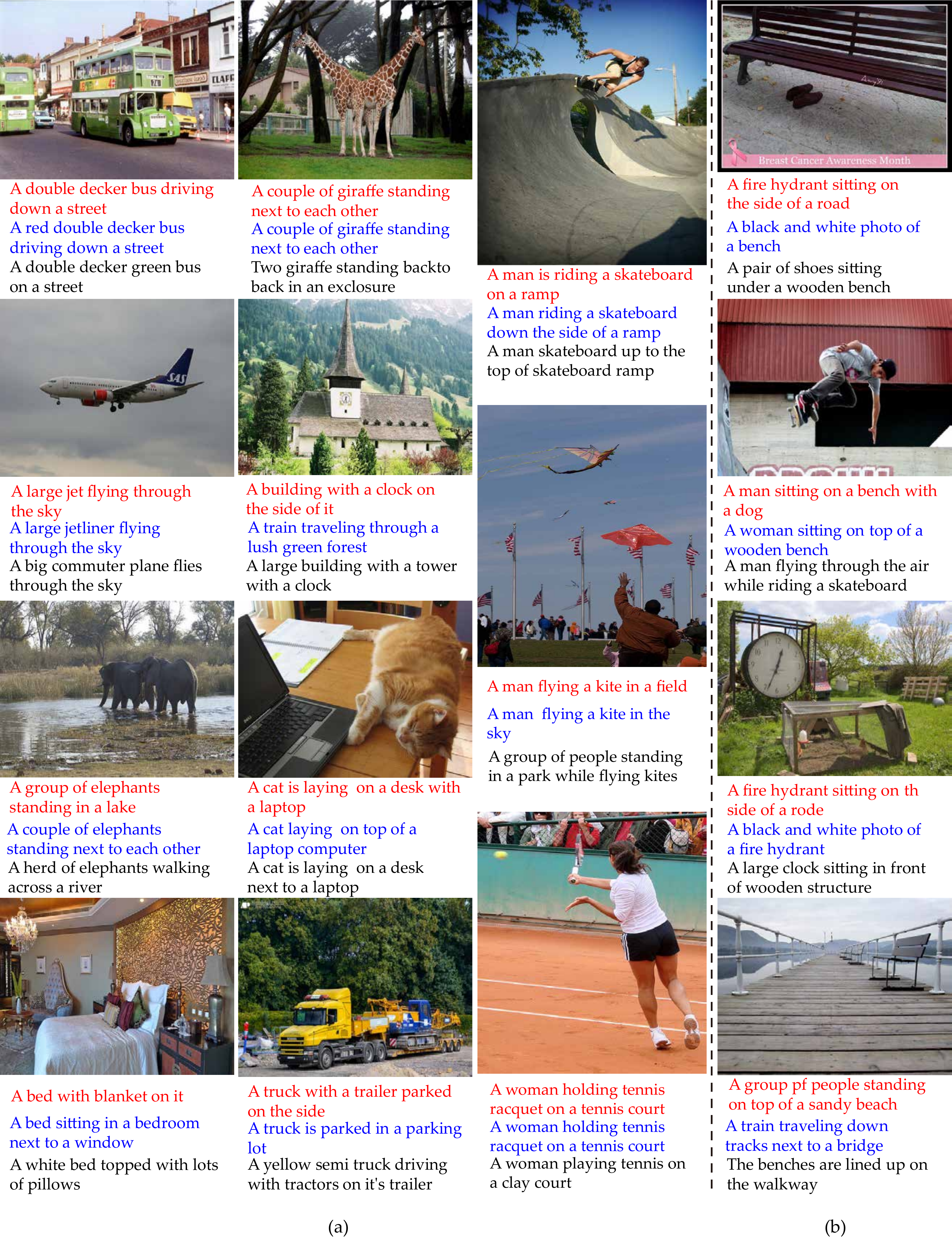}
  \caption{Sample generated descriptions on the MS COCO dataset. The sentences with \textcolor[rgb]{1.00,0.00,0.00}{red}, \textcolor[rgb]{0.00,0.00,1.00}{blue} and \textcolor[rgb]{0.00,0.00,0.00}{black} font are generated by our m-GRU model, DeVS \cite{karpathy2015devs} + LSTM model and human, respectively. The last column shows some failure cases.}\label{fig.5}
\end{figure*}
\par  Some visualized captioning results on Flickr8K and Flickr30K in Fig. \ref{flickr}. The results show that our model not only can describe the main content very well but also can generate grammatical correct sentences. Some results for sentence generation on the MS COCO dataset are also shown in Fig. \ref{fig.5}. The blue font sentences are generated by DeVS \cite{karpathy2015devs} but the villa RNN is substituted by the LSTM model because the LSTM model is more powerful than the vanilla RNN model. Despite doing this, the ``DeVS + LSTM'' model is less powerful than our m-GRU model. For example, in the third image of the second column, the sentence generated by our m-GRU model is similar with the sentence generated by human, but the sentence generated by ``DeVS + LSTM'' has something wrong, the cat is laying next to the laptop but ``DeVS + LSTM'' thinks the cat is on the laptop. Some failed cases are also be showed in Fig. \ref{fig.5}(b). We think the reason leads to generating wrong sentences is that the scene is too complex to recognize objects in it. Some objects may be not recognized by human at the first glance such as the second image of the forth column.

\par Through analyzing the results of experiments on the three benchmark datasets, the ``CNN + RNN'' model shows a strong ability to solve the image description problem and the original methods has been greatly improved by our m-GRU model. The results confirm the truth that the GRU network is good at long-term memory like the LSTM network but much simpler than LSTM which makes our model much easier to train, this is also very important for large scale data. Another discovery from the experiments results is that with the dataset's scale increasing, our performance becomes better and better.

\par Data transfer learning also researched in this paper. When we use the model trained on Flick30K to test on Flickr8K, the B-1 scores has gained 4 points more than the model trained on Flickr8K. This shows that with the training set size increasing, the performance of the model will be improved clearly. However, the same phenomenon does not appear when we use the model trained on MS COCO to test on Flickr8K. On the contrary, the B-1 scores degrade. This is because Flickr30K is the augmentation of Flickr8K, but the collection of MS COCO are different. In other words, images and sentences in Flickr8K are similar to Flickr30K, but are different from MS COCO.
%

\section{Conclusion} \label{conclusion}
In this paper, we propose a multi-modal GRU model for image description. This model is an end-to-end neural network model that can automatically generate an English sentence to describe the image. Our model uses CNN to encode the input image into an vector representation, then utilities a followed GRU network to generate a correlated sentence. The experiments on the three benchmark datasets show that the proposed model especially the 2-layer model can complete the automatical image description task appropriately. With the increasing size of dataset, our multi-modal GRU model can show a better performance for image description task.

\begin{acknowledgements}
This work was supported in part by the National Natural Science Foundation of China under Grant 61761130079, in part by the Key Research Program of Frontier Sciences, CAS under Grant QYZDY-SSW-JSC044, in part by the National Natural Science Foundation of China under Grant 61472413, in part by the National Natural Science Foundation of China under Grant 61772510, and in part by the Young Top-notch Talent Program of Chinese Academy of Sciences under Grant QYZDB-SSW-JSC015.
\end{acknowledgements}

\bibliographystyle{spmpsci}      
\bibliography{trans_cgs}   

\end{document}